\documentclass[runningheads]{llncs}
\usepackage[T1]{fontenc}
\usepackage{graphicx}
\usepackage{cite}
\graphicspath{{Figs/}}
\DeclareGraphicsExtensions{.pdf,.jpeg,.png}
\usepackage{xcolor}

\begin{document}
\title{Burnt area extraction from high-resolution satellite images based on anomaly detection}
\titlerunning{Burnt area extraction based on anomaly detection}
%
\author{Oscar David Rafael Narvaez Luces\inst{ 1,2}\and
Minh-Tan Pham\inst{ 1}\orcidID{0000-0003-0266-767X} \and
Quentin Poterek \inst{ 2}\orcidID{0000-0002-7666-6963}
\and
Rémi Braun \inst{ 2}\orcidID{0000-0002-5063-2087}}
\authorrunning{O. Narvaez Luces et al.}
%
\institute{IRISA, Université Bretagne Sud, UMR 6074, 56000 Vannes, France \and
ICube-SERTIT, Université de Strasbourg, 67412 Illkirch Graffenstaden, France\\
\email{minh-tan.pham@irisa.fr, remi.braun@unistra.fr}}
\maketitle              
\begin{abstract}
Wildfire detection using satellite images is a widely studied task in remote sensing with many applications to fire delineation and mapping. Recently, deep learning methods have become a scalable solution to automate this task, especially in the field of unsupervised learning where no training data is available. This is particularly important in the context of emergency risk monitoring where fast and effective detection is needed, generally based on high-resolution satellite data. Among various approaches, Anomaly Detection (AD) appears to be highly potential thanks to its broad applications in computer vision, medical imaging, as well as remote sensing.
In this work, we build upon the framework of Vector Quantized Variational Autoencoder (VQ-VAE), a popular reconstruction-based AD method with discrete latent spaces, to perform unsupervised burnt area extraction. We integrate VQ-VAE into an end-to-end framework with an intensive post-processing step using dedicated vegetation, water and brightness indexes. Our experiments conducted on high-resolution SPOT-6/7 images provide promising results of the proposed technique, showing its high potential in future research on unsupervised burnt area extraction.

\end{abstract}

\section{Introduction}
\label{sec:intro}




Wildfires are destructive events that involve a growing concern due to their devastating impact on ecosystems, infrastructures and even human lives \cite{fao2020global}. Early detection and mapping of wildfires play a crucial role in effective fire management and mitigation efforts. Remote sensing using satellite imagery has become a precious and standard tool for monitoring and analyzing wildfire events. Thanks to the ability to periodically capture high and very high spatial resolution images with multispectral bands, satellite sensors could provide a large amount of data allowing a large-scale observation and comprehensive view of fire incidents \cite{franklin2010remote}. These images offer valuable information about fire dynamics, burned area extent, and smoke dispersion patterns, helping to assess fire severity and facilitating timely decision-making. Therefore, there is a crucial and eventually urgent need for wildfire detection using satellite remote sensing images, allowing automatic and fast mapping of these events.


In recent years, deep learning has demonstrated remarkable success in various computer vision tasks, including object detection, classification, and segmentation. It has also emerged as a powerful tool in the field of remote sensing, revolutionizing the analysis and interpretation of remotely sensed data \cite{ma2019deep,yuan2020deep}. 
In the context of wildfire detection, deep learning techniques have shown immense potential in automating the identification and classification of fire non-fire regions accurately, by training these networks on labeled images using a fully supervised approach \cite{florath2022supervised}. However, most supervised deep learning methods have been still suffering from a main challenge of label availability to train their networks, due to the anomalous occurrence of wildfires \cite{seydi2022burnt, farasin2019unsupervised}. As a result, unsupervised anomaly detection (AD) \cite{pang2021deep} becomes a promising approach to fill the gap in using deep learning to perform burnt area detection from satellite images. Such an approach does not suffer from the unavailability of burnt labels (as their nature), as the training process is performed only using normal image data, which are available in huge amounts since most acquisitions of satellite sensors provide normal non-burnt scenes. Furthermore, during the prediction phase, a well-trained AD model could be able to detect burnt areas as anomalies.


Literature studies have shown the benefits of unsupervised deep neural networks for detecting burnt areas in visual spectral bands (i.e., RGB). In \cite{farasin2019unsupervised}, the authors proposed a multi-temporal approach with image data between the fire's starting and ending dates. They performed hue-difference segmentation and color-based segmentation techniques using both RGB and HSV color spaces. This approach is limited since we need both the \emph{pre-event} and \emph{post-event} images to perform the dedicated framework. 
In the scope of using mono-temporal images under an unsupervised strategy, the authors in \cite{coca2021anomaly} leveraged the autoencoder (AE) model to perform an outlier detection method based on the one-class support vector machines (OCSVM) to estimate burnt coverage in Sentinel-2 images. However,  challenges have been observed in the applicability of this algorithm across different regions due to variations in surface reflectance, backscatter coefficients, and seasonal changes \cite{coca2021anomaly}. Recently, a self-supervised method based on a Dirichlet distribution has been investigated in \cite{coca2022hybrid}. The authors have shown promising results on Sentinel-2 images by effectively limiting false positives, primarily caused by cloud and smoke presence in the images \cite{coca2022hybrid}. Similarly, a self-supervised contrastive learning technique based on SimCLR \cite{chen2020simple} has been employed to extract features from high-resolution data for wildfire delineation in \cite{zhang2022unsupervised}. Such a learning approach has demonstrated its effectiveness in identifying and segmenting wildfire regions with the absence of labeled data. 

In this paper, we draw the focus on burnt detection from high-resolution satellite images using an anomaly detection approach. We build upon the recent work \cite{gangloff2022leveraging} of AD using Vector Quantized Variational Autoencoder (VQ-VAE), a popular extension of VAE with discrete latent spaces. Indeed, several studies have shown the superior performance on reconstruction ability of VQ-VAE over the standard VAE in the computer vision domain \cite{marimont2021anomaly, gangloff2022leveraging} as well as in aerial remote sensing \cite{pham2023weakly}. Here, we investigate its capacity for burnt detection using high-resolution satellite images. We also propose an intensive post-processing step using several vegetation, water and brightness indexes, to benchmark the performance of our developed framework using two large-scale SPOT-6/7 images. 

The rest of this paper is organized as follows. In Section \ref{sec:data}, we describe our studied datasets with high-resolution SPOT-6/7 images from two different locations. Section \ref{sec:method} then describes the framework of VQ-VAE for burnt area extraction as well as the dedicated post-processing developed in our work. In Section \ref{sec:exp}, experimental results on two different scenarios are provided, quantitatively and qualitatively. Section \ref{sec:conclusion} finally concludes our paper and provides some perspective works for future studies on the topic.

\section{Datasets}
\label{sec:data}

\begin{figure}[htb!]
    \centering
    \includegraphics[width=0.47\textwidth]{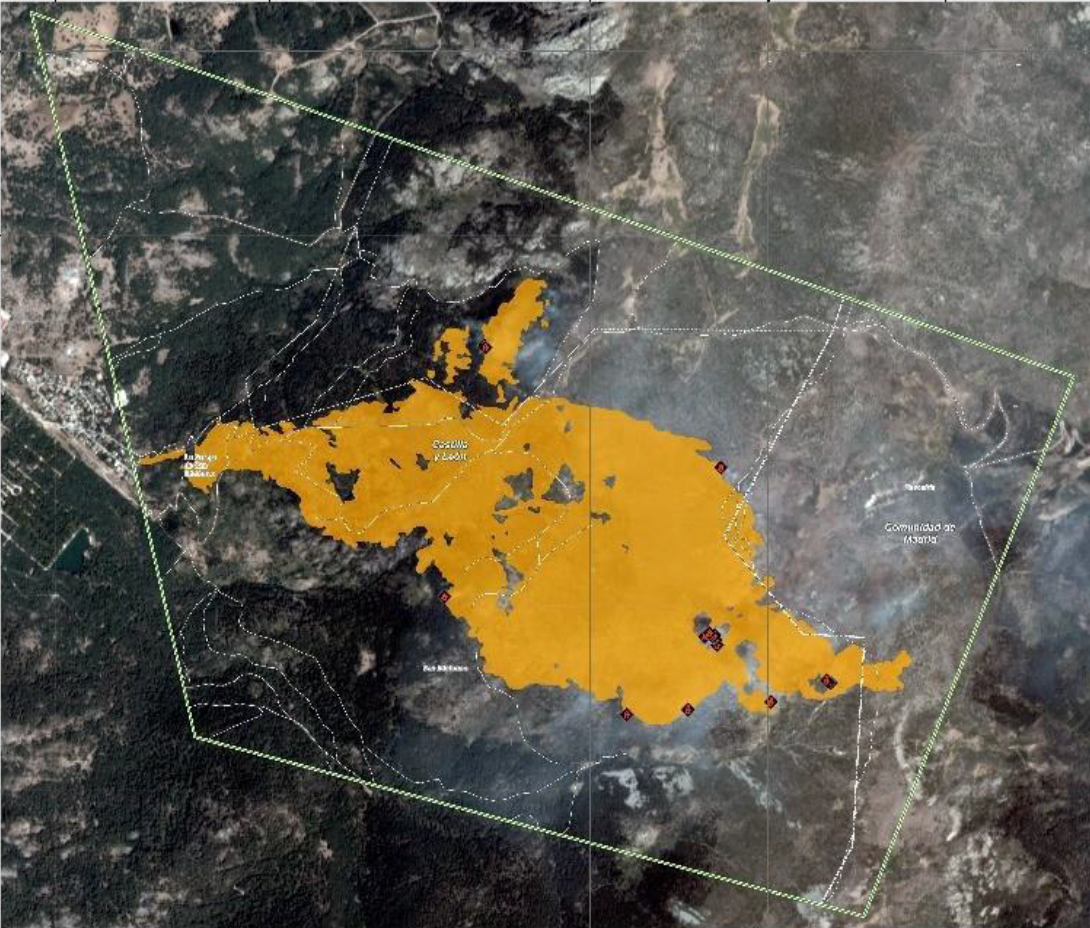}
    \hfill
    \includegraphics[width=0.47\textwidth]{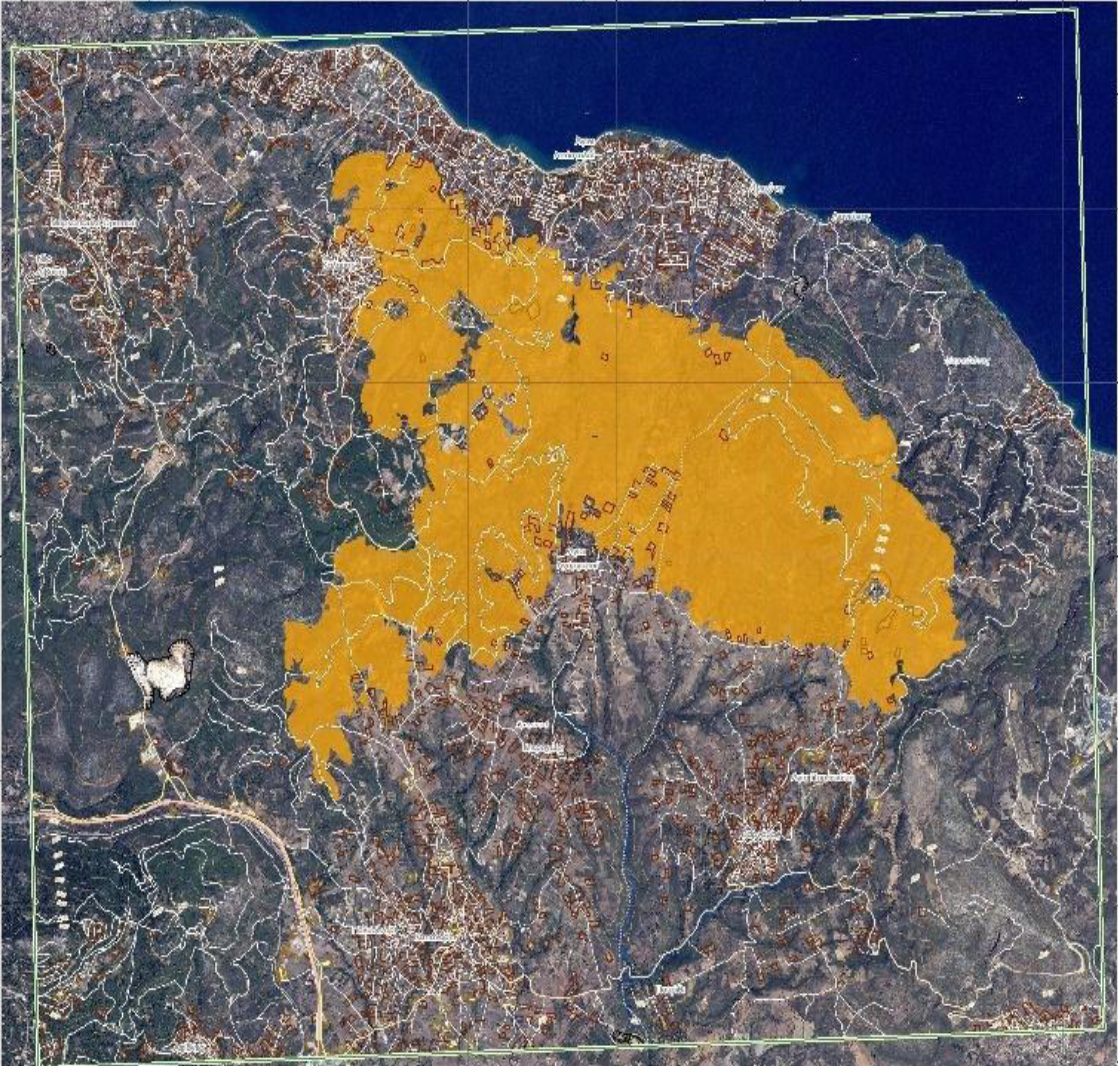}
    \caption{Illustration of delineation maps for the fire activation EMSR375 in Segovia, Spain (left) and the EMSR224 in Kalamos, Greece (right) \cite{sertit_emsr375_2019, sertit_emsr224_2017}, \emph{Copernicus Emergency Management Service} (\copyright European Union, 2012-2023). We note that the \emph{pre-event} image of EMSR375 will be used for training; the \emph{post-event} images of EMSR375 and EMRS224 will be used in prediction in our experimental study.}
    \label{fig:data}
\end{figure}

One of the important tasks in this work is the creation of the studied datasets from high-resolution satellite images dedicated to the topic of burnt area extraction. 
This was done based on the retrieval of metadata information
describing the historical summary of wildfire events from 2016 to 2018 thanks to the
Emergency Management System service of the Copernicus Program (CEMS). The data have been handled with EOReader, an open-source Python library for remote sensing developed by the ICube-SERTIT platform (Regional image processing and remote sensing service) \cite{maxant2022extracteo}. 
The following investigations were done to select datasets for our study:
\begin{itemize}
  \item Analysis of satellite sensors to focus the study. Different sensors are used for fire activation, including Sentinel-2, Pléiades, SPOT-6/7, GeoEye, etc.  
  \item Availability of the \emph{pre-event} and \emph{post-event} images with the same sensor (or different sensors at close spatial resolutions). This is particularly important since \emph{pre-event} and \emph{post-event} images are often acquired from different sensors. For the sake of simplicity, we aimed to conduct our work on image data at a single (or close) spatial resolution.
  \item Analysis of the complexity of the area in terms of diversity of land, presence of urban areas and cloud coverage. The objective was to perform and evaluate burnt area extraction from different land-cover scenes, e.g. both forest and urban zones.
\end{itemize}
As a result of those analyses, two fire activations were selected (and illustrated in Fig. \ref{fig:data}):
\begin{itemize}
  \item EMSR375 event in Segovia, Spain (August 5, 2019) \cite{sertit_emsr375_2019} with SPOT-6 images. This represents the study scenario of wildfires in forest areas with a low presence of urban areas.
  \item EMSR224 event in Kalamos, Greece (August 18, 2017) \cite{sertit_emsr224_2017} with SPOT-7 images. This represents a more complex scenario with a burnt areas surrounded by more extended urban area.
\end{itemize}
Next, two following scenarios were proposed to form our final datasets:
\begin{itemize}
  \item \textbf{Scenario 1:} train with the \emph{pre-event} image and test on the \emph{post-event} image, both of EMSR375. This is a usual and simple scenario when the training and testing are performed on the same scene (with a wildfire activation from forest zones, Fig. \ref{fig:data}) at the same geolocation.
  \item \textbf{Scenario 2:} train with the \emph{pre-event} image of EMSR375 and test on the \emph{post-event} image EMSR224. This is a more complicated scenario since the evaluation will be performed on another scene (with a wildfire activation surrounded by some urban zones, Fig. \ref{fig:data}) from another geolocation.
\end{itemize}

Three spectral bands including near-infrared, red and green were used to create false color images. We note that our approach could work with any band and any number of bands. Here, we chose to exploit those three bands without loss of generality.
To create the training set, the \emph{pre-event} EMRS375 image 
was patched into $256\times 256$ pixels with a stride of $128$ pixels, which resulted in a training dataset with 1654 patches. 
A pre-processing step was applied based on vertical and horizontal flipping, random rotations and a Gaussian blur filter (with a kernel of $11\times 11$ and a sigma of $5$) to reduce the effect of fine man-made structures, such as roads or urban areas.

\section{Methodology}
\label{sec:method}
\begin{figure}[htb!]
    \centering
    \includegraphics[width=0.99\textwidth]{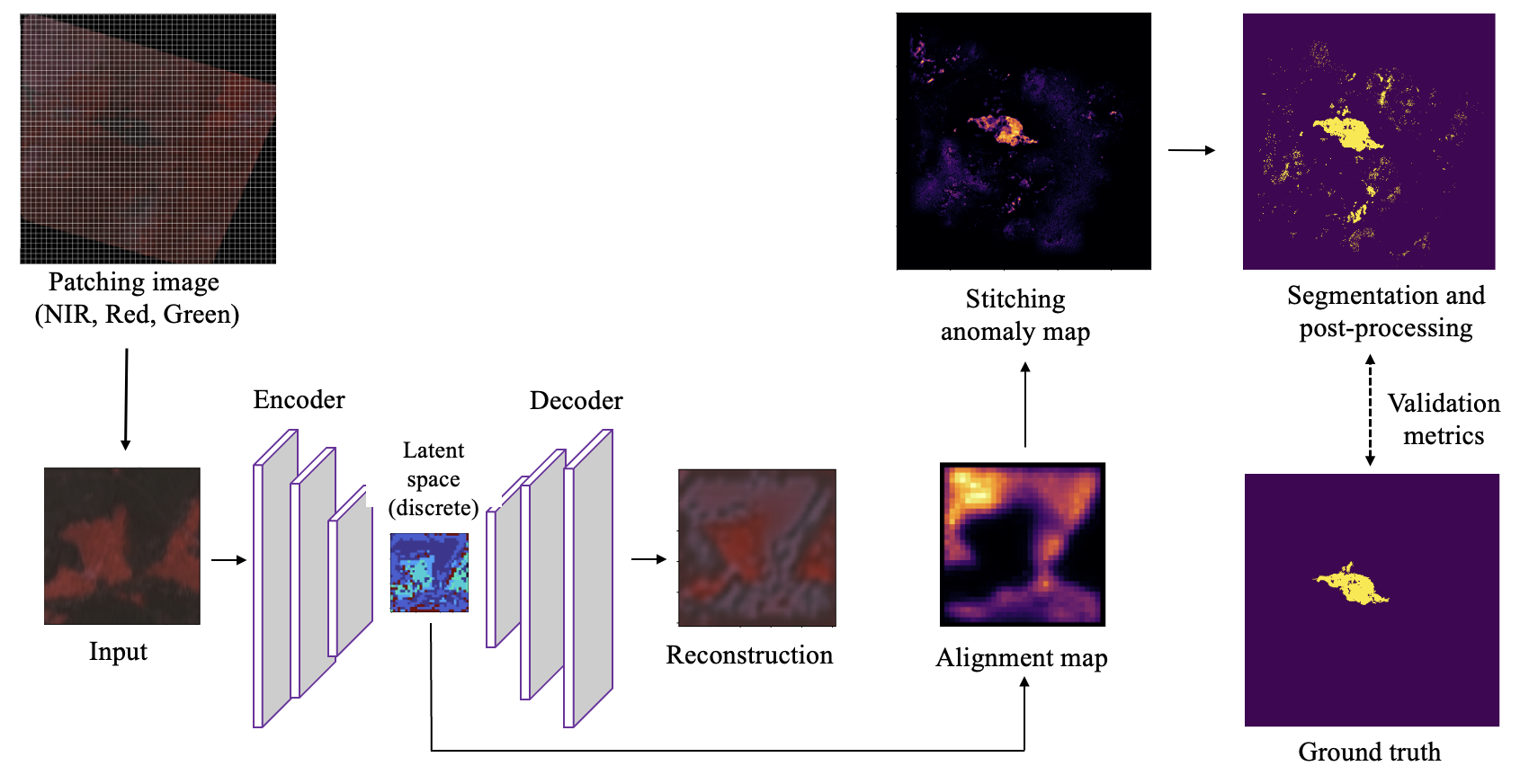}
    \caption{Overall framework for burnt area extraction from satellite image based on anomaly detection. Here, the final anomaly score is derived based on the alignment map computed on the discrete latent space of the VQ-VAE.}
    \label{fig:workflow}
\end{figure}

Our developed framework for burnt area extraction from satellite images is summarized in Fig. \ref{fig:workflow}. We build upon the work of VQ-VAE for anomaly detection in \cite{gangloff2022leveraging} and integrate it into the whole workflow with pre-processing and post-processing components. We first summarize the general idea of VQ-VAE for anomaly detection and its application in our study, then provide details on the important post-processing step.

\subsection{Anomaly detection using VQ-VAE}
Over the past few years, VQ-VAE with discrete latent variables \cite{van2017neural} has appeared to be one of the most popular extensions of the standard VAE model thanks to its capacity to provide high-quality reconstructions. Within the context of anomaly detection, the authors in \cite{gangloff2022leveraging} introduced a novel inner metric, namely \emph{alignment metric}, which was computed from the discrete latent space. To train the VQ-VAE model, this alignment term was added to the original two-term loss function (including the reconstruction and regularization terms) of VQ-VAE, as follows:
\begin{equation}
    \mathcal{L}^{VQ-VAE}_{total} = \mathcal{L}_{reconstruction} + \mathcal{L}_{regulization} + \mathcal{L}_{alignment}
\end{equation}

In the prediction phase, such an alignment map (AM) was also combined with the classical reconstruction-based anomaly map (denoted by SM in \cite{gangloff2022leveraging}) to yield the final anomaly score. This combination has improved the performance of anomaly detection in several vision datasets.
For more details about the computations and fusion of these two metrics, we refer readers to \cite{gangloff2022leveraging}. In our work, it should be noted that only using AM to measure anomaly score provides better performance than the combination with SM. Thus, we finally discard the SM and only keep the AM as our anomaly score for burnt areas. We will discuss about this behavior later in our experimental study and conclusion.

\subsection{Post-processing}

In order to produce the final anomaly map for the entire image tile, post-processing step becomes crucial and should be done carefully. As illustrated in Fig. \ref{fig:workflow}, our post-processing involves three steps:
\begin{enumerate}
    \item Stitching anomaly maps from patches into the entire large-size map;
    \item Performing binary segmentation to obtain the binary anomaly map;
    \item Applying some thresholds based on vegetation, water and brightness indexes to remove false positives (i.e. non-burnt areas detected as anomalies). In detail, the following indexes were used for the final thresholding step: 
\begin{itemize}
    \item[+] Normalized Difference Vegetation Index (NDVI)
    
    \begin{equation}
        NDVI = \frac{NIR-Red}{NIR+Red}
    \end{equation}
    
    \item[+] Normalized Difference Water Index (NDWI)
    \begin{equation}
        NDWI = \frac{Green-NIR}{Green+NIR}
    \end{equation}
    \item[+] Landsat-based Brightness Index (TMBI)
    \cite{mathieu1998BITM}
    \begin{equation}
        TMBI = \sqrt{\frac{Blue^2+Green^2+Red^2}{3}}
    \end{equation}
\end{itemize}
\end{enumerate}




It should be noted the the threshold values for these indexes could be set using expert knowledge and literature regarding burnt area detection. However, in practice, one could automatically select the most suitable values based on the precision/recall analysis maximizing the F1-score for each index.

\section{Experimental study}
\label{sec:exp}

\subsection{Model and training setup}
We exploit the same model architecture as in \cite{gangloff2022leveraging} with 3 convolutional layers and an input image size of $256\times 256$. The dimension of the latent variable is 32 and the number of embedded vectors is 256. Learning rate and batch size are set to $10^{-4}$ and $16$, respectively. 

We trained the model with $200$ epochs on the training set including only non-burnt patches (i.e., normal data) extracted from the \emph{pre-event} EMSR375 image, as described in Section \ref{sec:data}. False color images using three spectral bands of near-infrared, red, green were used as the network input. Nevertheless, as previously mentioned in Sec. \ref{sec:data}, the model can be trained using all 4 bands, three bands of RGB, well as well the pan-sharpened version of SPOT-6/7 images without any issue since it is trained from scratch. 


\subsection{Results and discussions}
Table \ref{table:res} provides the Recall, Precision and F1-score yielded by the developed model on the two studied scenarios, before and after applying post-processing. From the table, we observe a clear improvement in precision (and thus in F1-score) by using post-processing to eliminate false positives. As a result, the final values of F1-score values were $0.588$ for Scenario 1 and $0.675$ for Scenario 2. These are  preliminary benchmarking results on our dedicated dataset, allowing future work to perform a comparative study.

Regarding the two scenarios, even though for the first one we performed prediction on the \emph{post-event} image at the same scene as the \emph{pre-event} EMRS375 for training, the performance is lower than in Scenario 2. This is because of the complexity of the landscape and more importantly, the presence of cloud in the \emph{post-event} image. Cloud appearance will indeed reduce the performance of anomaly detection, since at some points, cloud could be considered as anomaly if it was not considered and learned as normal data during training. Future works should be focused on this issue by using a robust cloud removal algorithm or performing a fusion of radar-optical data. Back to our analysis on Scenario 2, we achieved a better performance although the prediction was performed on the \emph{post-event} of another image (EMSR224) than the one used for training. This shows the highly scalable capacity of the proposed approach. Training enough normal data across different geolocations could make the model relevant to extract burnt areas at a global scale. Yet, seasonal and sensor changes are about to be other aspects to be considered in future studies.

\begin{table}[ht]
\centering
\begin{tabular}{l||c|c||c|c}
 \hline
 & \multicolumn{2}{c||}{\bf Scenario 1}  & \multicolumn{2}{c}{\bf Scenario 2}   \\
 \hline
 & No post-proc.  & With post-proc. & No post-proc.  & With post-proc.  \\

 \hline
Recall & 0.836 & 0.830 & 0.591 & 0.692\\
Precision & 0.214 & 0.455 & 0.590 & 0.660\\
F1-score & 0.341 & {\bf 0.588} & 0.591& {\bf 0.675} \\
 \hline
\end{tabular}
\caption{Quantitative results on the two scenarios without and with the post-processing step. Our model is trained on the \emph{pre-event} of the EMSR375. Burnt areas are extracted from the \emph{post-event} of the EMSR375 \textbf{(Scenario 1)} and the post-event of the EMSR224 \textbf{(Scenario 2)}.}
\label{table:res}
\end{table}

\begin{figure}[htb!]
    \centering
    \includegraphics[width=0.6\textwidth]{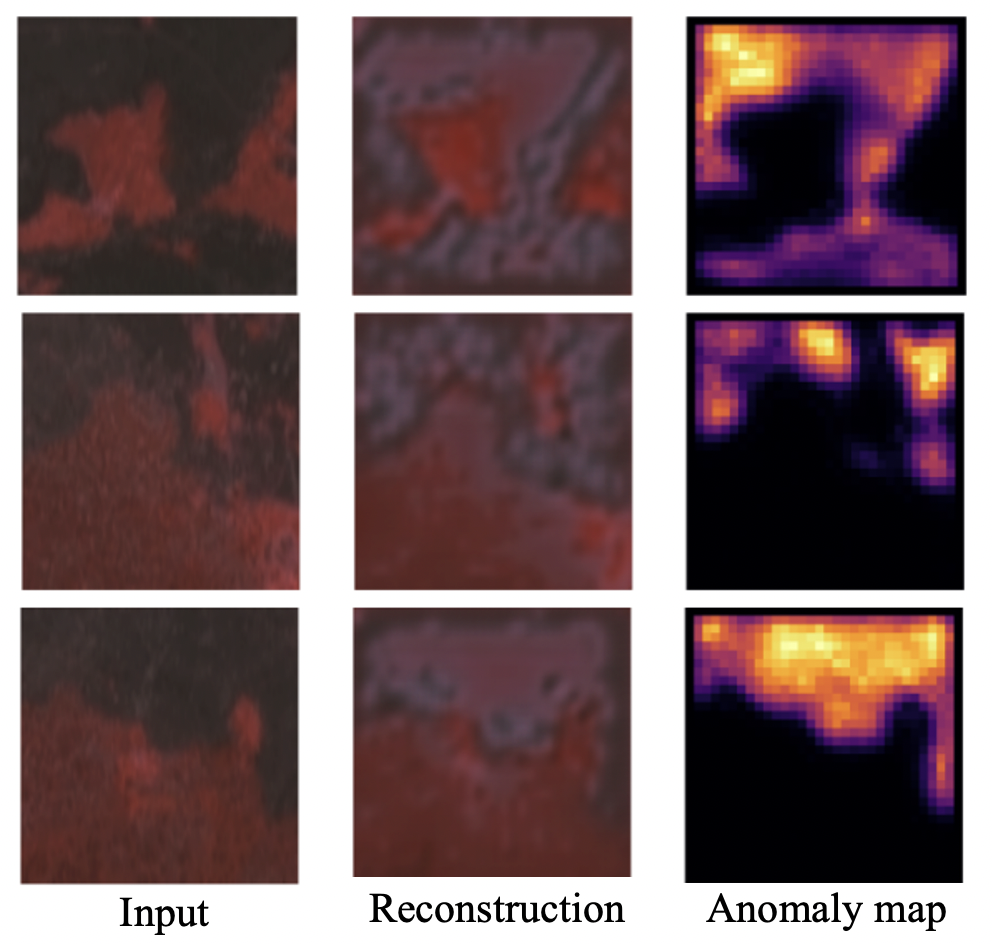}
    \caption{Visual illustration of model reconstructions. Burnt areas (seen as black from input images) could not be reconstructed by the VQ-VAE since it was trained only on non-burnt images. This allows to yield good-quality anomaly map.}
    \label{fig:result}
\end{figure}

In order to provide some qualitative assessment of the developed method, we illustrate the reconstruction outputs and the anomaly maps yielded for some inputs in Fig. \ref{fig:result}. Here, we depict sample inputs that could help readers to understand the mechanism of the reconstruction-based AD approach. As observed in the figure, the VQ-VAE model was not able to reconstruct burnt areas (black zones in input images) since it had been only trained on normal non-burnt data (i.e. from the \emph{pre-event} image). The model attempted to reconstruct those regions as best as it can, yielding highly structured motifs which help us to easily recognize anomalies in reconstructed scenes. Although our method does not exploit the direct reconstruction-based anomaly map, this observation still indirectly shows the quality of VQ-VAE latent spaces to encode and reconstruct image features. To this end, the quality of the alignment map computed from the latent space has a strong coherency with the reconstruction capacity.

\section{Conclusion}
\label{sec:conclusion}
This work has shown a high potential of unsupervised anomaly detection to extract burnt areas from high-resolution satellite images. Our preliminary results on the two studied scenarios may provide the first ingredient to perform a 
comparative study to other approaches in future work.

During our experimental study, several observations were found related to the behavior of the VQ-VAE model in the scope of burnt anomaly detection. Firstly, the alignment map AM is a metric not directly related to reconstruction output but from the discrete latent space, leading to the idea that it might not be the best metric to delineate the anomalies. However, our study has shown that the only use of AM provided better performance than the combination of AM and the reconstruction-based metric SM proposed in \cite{gangloff2022leveraging}. Such a find shows that this inner metric has a huge potential in anomaly detection. It might depend on the nature of anomalies to choose the best role of AM in the final anomaly map. This aspect should be also investigated in future works. 



Secondly, the application of the model trained on a small dataset (for example, a single SPOT-6 image as in our experiment), then applied to different geolocations still remains a challenge due to variations in land-cover scenes. Objects with high reflectance which have not been seen by the model during training can be easily identified as anomalies in prediction. Fortunately, the dedicated post-processing steps seem to handle well the second scenario in our experiment. Nevertheless, this remains a crucial step that should be further investigated to better deal with several different types of outliers.

Regarding the perspective works, besides the investigation of anomaly metrics in the latent space, other recent VAE-based models for AD such as \cite{gangloff2022variational} could be explored. Also, one may be interested in using self-supervised approaches \cite{berg2022self} to foster the representative ability of VAE-like models.



%
%
%
\bibliographystyle{splncs04}
\bibliography{refs}

\end{document}